\documentclass[journal]{IEEEtran}
\usepackage[utf8]{inputenc}
\usepackage{amsmath}
\usepackage{graphicx}
\usepackage{float}
\usepackage{amssymb}
\usepackage[noadjust]{cite}
\usepackage[font=footnotesize]{caption}
\usepackage{subcaption}
\usepackage{cleveref}
\usepackage{placeins}

\begin{document}

\title{High-Speed CMOS-Free Purely Spintronic Asynchronous Recurrent Neural Network}
\author{Pranav O. Mathews, \IEEEmembership{Graduate Student Member, IEEE}, Christian B. Duffee, \IEEEmembership{Graduate Student Member, IEEE}, Abel Thayil, Ty E. Stovall, Christopher H. Bennett, \IEEEmembership{Member, IEEE}, Felipe Garcia-Sanchez, \IEEEmembership{Member, IEEE}, Matthew J. Marinella, \IEEEmembership{Senior Member, IEEE}, Jean Anne C. Incorvia, \IEEEmembership{Member, IEEE}, Naimul Hassan, \IEEEmembership{Graduate Student Member, IEEE}, Xuan Hu, \IEEEmembership{Graduate Student Member, IEEE}, Joseph S. Friedman, \IEEEmembership{Senior Member, IEEE\vspace{-1.5em}}

\thanks{Manuscript submitted XXXX XX, 2022. This research is sponsored in part by the National Science Foundation under CCF award 1910800 and the Texas Analog Center of Excellence undergraduate internship program. \textit{(Pranav O. Mathews, Christian B. Duffee, and Abel Thayil contributed equally to this work)} \textit{(Corresponding author: Xuan Hu)}}
\thanks{P. O. Mathews is with the School of Electrical and Computer Engineering, Georgia Tech, Atlanta, GA 30332, USA and the Department of Electrical and Computer Engineering, The University of Texas at Dallas, Richardson, TX 75080, USA}
\thanks{C. B. Duffee is with the Department of Electrical and Computer Engineering, Northwestern University, Evanston, IL 60208, USA and the Department of Electrical and Computer Engineering, The University of Texas at Dallas, Richardson, TX 75080, USA}
\thanks{A. Thayil is with the Laboratoire de Physique de la Matière Condensée, Ecole Polytechnique, CNRS, IP Paris, Palaiseau 91128, France and the Department of Electrical and Computer Engineering, The University of Texas at Dallas, Richardson, TX 75080, USA}
\thanks{T. E. Stovall, N. Hassan, X. Hu, and J. S. Friedman are with the Department of Electrical and Computer Engineering, The University of Texas at Dallas, Richardson, TX 75080, USA (email: joseph.friedman@utdallas.edu).}
\thanks{C. Bennett is with Sandia National Laboratories, Albuquerque, NM 87185, USA}
\thanks{F. Garcia-Sahnchez is with Universidad de Salamanca, Departamento de Física Aplicada, Salamanca, 37008, Spain }
\thanks{M. Marinella is with the School of Electrical, Computer and Energy Engineering, Arizona State University, Tempe, AZ 85287, USA and Sandia National Laboratories, Albuquerque, NM 87185, USA}
\thanks{J. A. C. Incorvia is with the Department of Electrical and Computer Engineering, The University of Texas at Austin, Austin, TX 78712, USA}

}

\maketitle

\begin{abstract}

The exceptional capabilities of the human brain provide inspiration for artificially intelligent hardware that mimics both the function and structure of neurobiology. In particular, the recent development of nanodevices with biomimetic characteristics promises to enable the development of neuromorphic architectures with exceptional computational efficiency. In this work, we propose biomimetic neurons comprised of domain wall-magnetic tunnel junctions that can be integrated into the first trainable CMOS-free recurrent neural network with biomimetic components. This paper demonstrates the computational effectiveness of this system for benchmark tasks, as well as its superior computational efficiency relative to alternative approaches for recurrent neural networks.

\end{abstract}

\section{Introduction}
Recurrent neural networks (RNN) are computationally-powerful structures that can solve complex tasks that are challenging for conventional von Neumann systems. Their unique problem-solving capabilities have led to their widespread use in areas such as image recognition, classification, and artificial intelligence. However, as data sets become larger and tasks become harder, the cost to run and train software RNNs increases to impractical levels \cite{Parberry1996ScalabilityProblem, Keuper2017DistributedScalability}.  \par

An answer to these issues is to redesign networks using analog circuits instead of the current software implementations that are inefficiently mapped to CMOS von Neumann hardware. Inspiration can be drawn from the brain: an analog machine that efficiently solves problems that are challenging for modern computers \cite{Balasubramanian2015HeterogeneityBrain}. Research shows that the brain accomplishes this through a diverse system of neurons woven together in a complex web of specialized bio-circuitry that cascades together to solve tasks \cite{Balasubramanian2015HeterogeneityBrain}. By emulating neuronal behavior and cytoarchitectonics, efficient circuits can be designed that bolster the capabilities of neural networks. \par

As conventional CMOS devices do not naturally mimic the physics of the brain, alternative switching elements have been proposed for neuromorphic computing based on emerging device technologies. Hysteretic devices such as memristors \cite{prezioso2015training, thakur2018large,li2018review,zhang2017artificial,wang2018fully,pedretti2017stochastic} and spintronic devices \cite{NeuroSpinReview_grollier2020neuromorphic, samsung_jung2022crossbar, siddiqui2019magnetic,Hassan2018MagneticInhibition,brigner2022domain, Brigner2019Graded-Anisotropy-InducedNeuron,Brigner2019Shape-basedNeuron,kurenkov2020neuromorphic} have received particular interest, as hysteresis is a critical component of neurobiological learning. RNNs have been proposed based on a hybrid of memristors and CMOS, \cite{wu2012synchronization,smagulova2018design,long2016reram,guo2015modeling,cauwenberghs1996analog,YiARXIV2021} and fully spintronic reservoir computers have also been designed \cite{Zhou2020ReservoirArrays, Mesaritakis2020Micro-ring-resonatorNetworks, Vandoorne2014ExperimentalChip, Ganguly2017ReservoirP-Bits, Wu2012SynchronizationNetworks, Cao2017Fixed-timeNetworks, Smagulova2018ACircuit, Liu2020Memristor-basedApplications, Li2019LongArrays, Guo2015ModelingCircuits, Cai2020Power-efficientNetworks, Bevi2020DesignMemristors, Duan2016Small-worldRecognition}; however, the hybrid systems utilize a significant amount of inefficient CMOS, detracting from the benefits of using emerging device technologies, while the spintronic reservoir utilizes no CMOS elements but does not provide the ability to train the weights.

This paper therefore proposes the first CMOS-free trainable RNN with biomimetic components. Leveraging the domain wall-magnetic tunnel junction (DW-MTJ) \cite{currivan2016logic,alamdar2021domain,liu2021domain,shibata2020linear,Hu2019SPICE-OnlyLogic}, this efficient and flexible RNN system is composed solely of spintronic devices and allows trained weights to be written into the system. This system mirrors the function, structure, and asynchronous nature of biological neurons, and is able to recover corrupted versions of patterns with high accuracy at higher speeds and with less power than alternative architectures.

\section{Background}

The direct implementation of RNNs in efficient hardware circuits requires the interconnection of hysteretic switching elements that can drive other hysteretic switching elements. Unlike CMOS transistors that do not store information in a non-volatile manner, the DW-MTJ can store a non-volatile state with a programmable domain wall position. This non-volatility enables low-energy computing in which power is only dissipated when retrieving or storing a state, in contrast to other devices that also need power to maintain a state. Moreover, these devices can be easily chained together, without the need of external circuitry to translate between differing output and input formats. These properties make them ideal building blocks for hardware RNNs, with neurons connected in a recurrent manner.

\subsection{Recurrent Neural Networks}

An RNN is a neural network architecture that retains a state that evolves over time in response to varying inputs \cite{Lipton2015ALearning}. These architectures are commonly used to solve tasks involving classification of spatial data or temporally related data; however, they suffer from high computational costs, especially as the number of neurons increases \cite{OrponenAnNetworks}. While actually consisting of a fixed number of interconnected neurons, the evolution of the network's state can be visualized as a directed graph of arbitrary length, with one time step separating each layer in the graph.  

\begin{figure}
\centering
\includegraphics[width=3in]{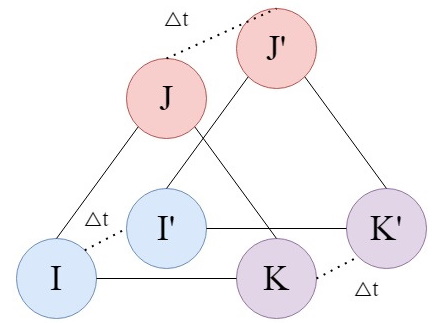}
\caption{Illustration of Hopfield network structure. Each triangle represents the interconnected neurons, and the dashed lines illustrate how the network steps over time, eventually converging to a final state. }
\label{Figure 1}
\end{figure}

The earliest example of an RNN is the Hopfield network \cite{Hopfield1982NeuralAbilities.}, shown in Fig. \ref{Figure 1}. Hopfield networks can store and remember patterns by functioning as associative memory, an important type of neural computation \cite{Yu2020AnNetwork}. However, because the network size scales with the data size \cite{OrponenAnNetworks} and the network is fully connected, excessive energy is required to calculate updates for large data sets.

A Hopfield network can be represented as a fully connected graph of neurons, where each neuron $i$ has an excitation state ($S_\textrm{i}$): ON (1) or OFF (0). Each neuron $i$ also sends an output ($O_\textrm{ij}$) to every other neuron $j$ in the network based on its current state and the synaptic weight ($W_\textrm{ij}$) between the pair of neurons 
\begin{equation}
    O_\textrm{ij} = S_\textrm{i} \times W_\textrm{ij}.
\end{equation}
As the network runs forward in time, each neuron’s state is updated based on its inputs according to the following expressions:
\begin{equation}
    S_i = 
    \begin{cases}
        1 \text{ if } \sum_{j=0}^{N-1} O_{ji} > \theta \\
        S_i \text{ if } \sum_{j=0}^{N-1} O_{ji} = \theta \\
        0 \text{ otherwise}
    \end{cases}
\end{equation}
where $\theta$ is some threshold value. The network can be trained to remember a certain pattern by updating the weights using the following equation:
\begin{equation}
    W_\textrm{ij} = (2S^{'}_\textrm{i} - 1)(2S^{'}_\textrm{j} - 1),
    \label{trainEq}
\end{equation}
where $S^{'}_i$ is the state of the $i$th bit of the training pattern, resulting in $W_\textrm{ij} = \pm 1.$

\begin{figure}
\centering
\includegraphics[width=3in]{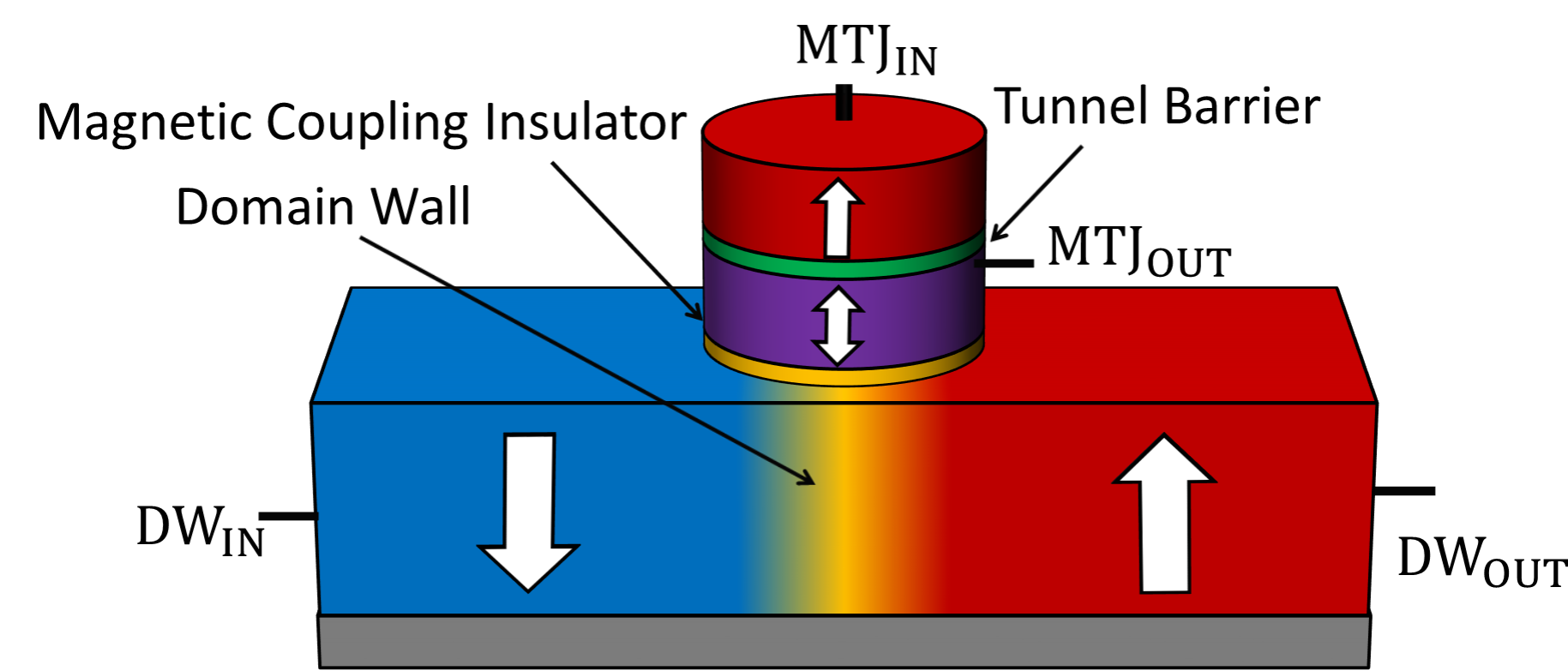}
\caption{Four terminal DW-MTJ with physical structure labeled. Input current flows between DW$_\textrm{IN}$ and DW$_\textrm{OUT}$, which can cause the DW to move. The free layer (purple) of the MTJ is coupled to the region of the DW track directly below it, and therefore switches when the DW passes under it. The MTJ resistance depends on the relative orientation of its two magnets, producing an output current through the MTJ between MTJ$_\textrm{IN}$ and MTJ$_\textrm{OUT}$.}
\label{Figure 2}
\end{figure}

\subsection{Four-Terminal Domain Wall-Magnetic Tunnel Junction}

The DW-MTJ is a spintronic device that has been used to implement non-volatile logic as well as neuronal and synaptic functions, offering improvements in efficiency and the ability to cascade into large-scale networks \cite{Brigner2019Graded-Anisotropy-InducedNeuron, Brigner2019Shape-basedNeuron, Brigner2020SPIN, Hassan2018MagneticInhibition, Brigner2020}. As shown in Fig. \ref{Figure 2}, a four-terminal DW-MTJ is composed of an MTJ atop a DW track separated by an electrically insulating layer \cite{Brigner2020}. 
The top ferromagnet's magnetization is fixed while the bottom ferromagnet is coupled with the DW track's magnetization directly beneath the MTJ. The MTJ resistance is larger when the two ferromagnets are in an anti-parallel state relative to a parallel state. Thus, the resistance of the MTJ is directly controlled by the DW's position in the track.\par

An input excitation such as a magnetic field or spin-transfer current can move the DW along its track length. When the DW passes the MTJ position, the MTJ free ferromagnet magnetically realigns itself with the track below, switching the MTJ between high and low resistive states. 

\section{Spintronic RNN with Biomemtic Neurons}

This paper proposes the first CMOS-free trainable RNN with biomimetic components, leveraging the hysteretic and analog properties of DW-MTJs \cite{Brigner2019Graded-Anisotropy-InducedNeuron, Brigner2019Shape-basedNeuron, Brigner2020SPIN, Hassan2018MagneticInhibition, Brigner2020} in a novel biomimetic neuron structure comprising multiple DW-MTJs. We further propose that recurrent networks can be realized by interconnecting synaptic outputs to dendritic inputs in fully-spintronic RNNs with high computational efficiency. 

\subsection{Biomimetic Neuron Structure \& Function}

\begin{figure}
\centering
\includegraphics[width=0.48\textwidth]{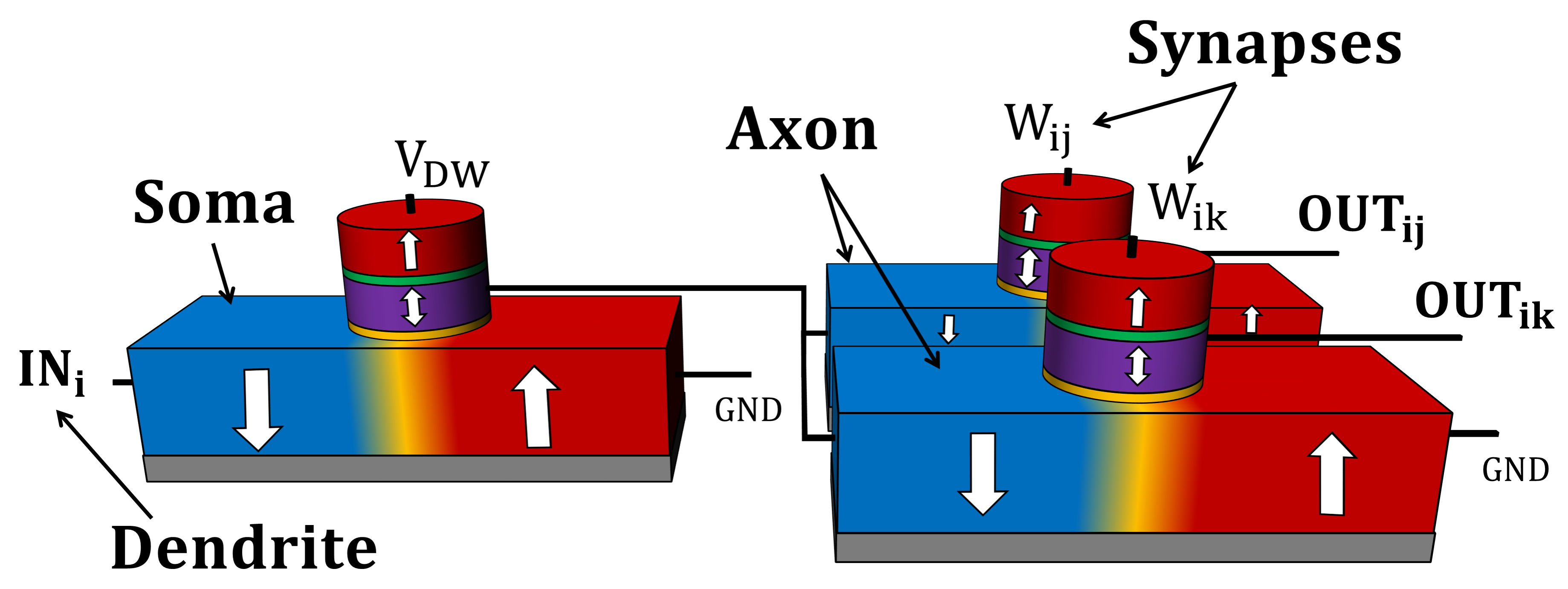}
\caption{A spintronic neuron with two axon branches. Input signals come through the dendrite, flow into the soma, and then into each axon. Each axon fires a synapse-weighted output of $\pm W_\textrm{ij}$ . }
\label{Figure 3}
\end{figure}

In neurobiological systems, a neuron continually stores and updates its state based on input excitations and transmits output signals as a function of that state; our biomimetic neuron is analogous both functionally and structurally to biological neurons, as illustrated in Fig. \ref{Figure 3}. The state of the neuron is represented in its body -- hereafter referred to as soma -- and updated by integrating the inputs it receives at its dendrite from pre-synaptic neurons. If the soma is sufficiently excited (\textit{i.e.}, its threshold is surpassed), a signal is fired to the several branches of the neuron's axon. The signal is modified by the synapses at the end of each axon, and propagated to the dendrites of post-synaptic neurons.\footnote{While the other terms closely match their biological counterparts in both form and function, \textit{soma} has been chosen here for lack of a more appropriate term.}

\begin{figure*}
\centering
\includegraphics[trim = 0 0 0 0 , clip, width=1\textwidth]{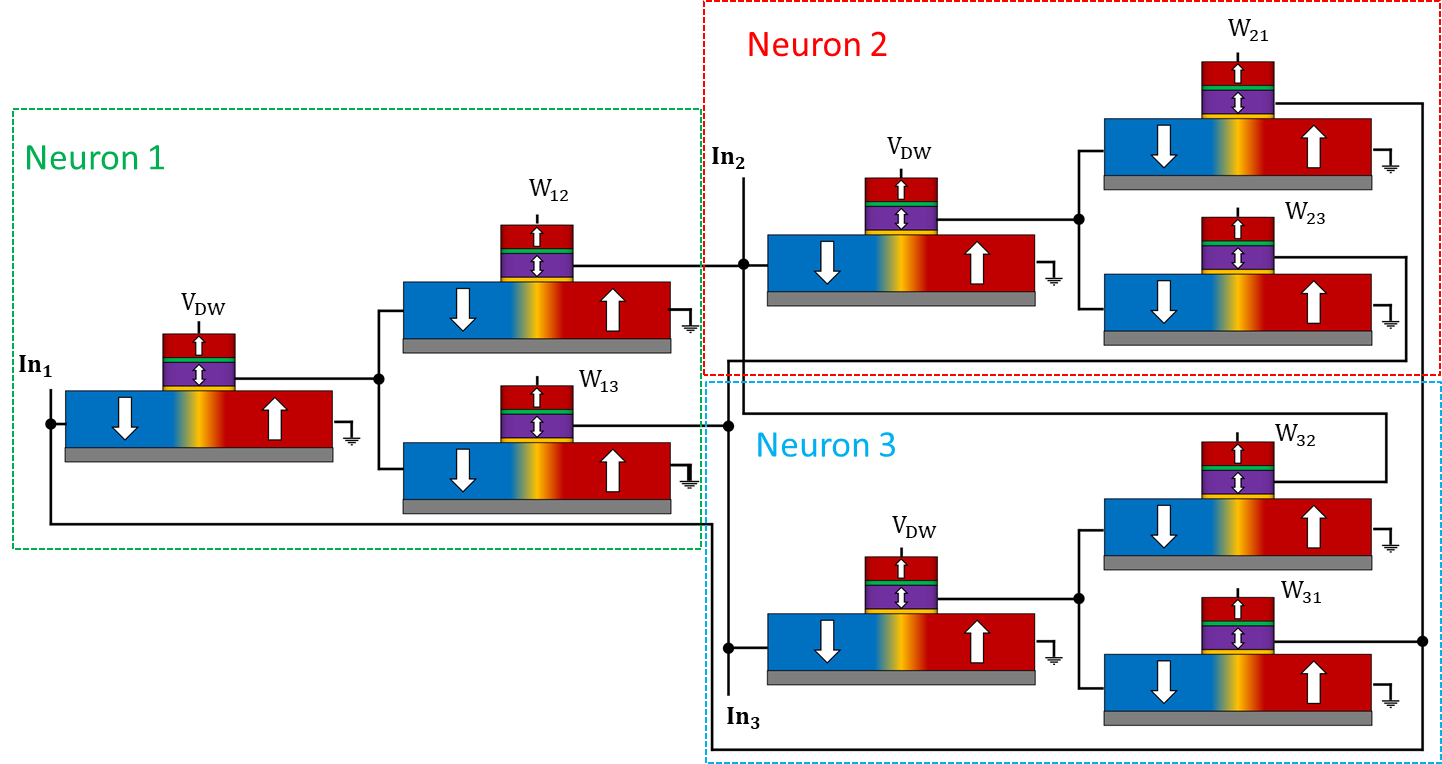}
\caption{Schematic of a three-neuron spintronic Hopfield network showing neuron interconnection. In$_\textrm{i}$ is the external input to neuron $i$, $V_\textrm{DW}$ is the soma-to-axon voltage, and $W_\textrm{ij}$ is the synaptic weight between neurons $i$ and $j$. The neurons are connected in an all-to-all scheme, affecting each other over time through their weighted connections and eventually converging to a stable state.}
\label{Figure 4}
\end{figure*}

\subsubsection{Dendrite}
In biological neurons, the dendrite is a branching structure that receives the synaptic output, represented as voltage differentials, of spatially distributed neurons \cite{HawkinsDendrite}. In our structure, the communication between neurons is represented as current levels. As such, the dendrite is an electrical node connected to each of the pre-synaptic neurons' synaptic outputs. 

\subsubsection{Soma}
In the proposed biomimetic neuron, the soma is a single DW-MTJ that integrates input from the neuron's dendrite and sends signals to its axons through its MTJ. The soma's excitation state is physically determined by the location of the DW with respect to the MTJ -- left of the MTJ corresponds to an off state, and right to an on state. Current through the DW track moves the DW in either direction, dependent on the direction of the input current. In an on state, the magnetization of the free and fixed ferromagnets are parallel across the MTJ, sending a high current signal to the axons. In an off state, the magnetic orientations are anti-parallel across the MTJ, sending a smaller signal and relaxing the axons.

When it is directly beneath the MTJ, the output signal is in an intermediate state between its high and low signals; to prevent the DW from staying in this intermediate state, the soma of the artificial neuron employs a leaking factor that relaxes the DW toward one edge of the track when no input is applied. Analogous to the method by which biological neurons maintain homeostasis with ion leakage channels to create a leaky integrate-and-fire system, our biomimetic neuron can perform leaking via a dipolar magnetic field, an anisotropy gradient, or a shape gradient \cite{Brigner2020SPIN, Brigner2019Graded-Anisotropy-InducedNeuron, Brigner2019Shape-basedNeuron}. This leaky integrate-and-fire system helps modulate the neuron's excitation when insufficiently stimulated and serves as a means of temporal coincidence detection between stimuli \cite{trappenberg_2010}. 

\subsubsection{Axons}
All axon branches are DW-MTJs connected to the soma's MTJ, as shown in Fig. \ref{Figure 3}. In order for a change in state to propagate between neurons, a soma's axons must mirror the soma's state. While a high current causes the soma to pull its axons into the parallel state, a low current does not push the axons' DWs towards the anti-parallel state. An appropriate threshold leaking force is used instead \cite{Brigner2019Graded-Anisotropy-InducedNeuron, Brigner2019Shape-basedNeuron, Brigner2020SPIN} that outweighs the low current signal, moving the axon's DW towards an anti-parallel state. In contrast, a high current signal is greater than the leaking force, and thus pushes the DW towards a high state. In a high state, the axon fully fires its weighted connection toward another neuron.

\subsubsection{Synapse}
The synaptic weight voltages on top of the axons function as the synapses, as these are where the connection weights are manipulated. A positive voltage value corresponds to an excitatory relation with the post-synaptic neuron, while a negative voltage value corresponds to an inhibitory relation. This is analogous to the brain, where information is stored not in the neurons themselves, but in the synaptic connections between them \cite{Kandel2001}. As such, the synaptic weight is represented by the MTJ voltage and is set externally during training.

\subsection{Spintronic RNN}

These biomimetic neurons can be directly interconnected to form an RNN, without requiring any CMOS or other circuitry. A simple RNN, the Hopfield network, is illustrated in Fig. \ref{Figure 4} with all-to-all connectivity. In this case of all-to-all connectivity, every neuron in an N-neuron network has one soma and N-1 axons. In total, an N-neuron Hopfield network consists of $\text{N}^2$ DW-MTJ devices.\par

Training is performed offline according to (\ref{trainEq}), with the resulting weights applied as synapse voltages of $\pm$W. For multiple patterns, the formula is applied individually for each and then the weight matrices are summed to form the final set. \par

To perform a computational task on a particular input, the network states can be initialized to an input pattern using voltage sources of constant \( \pm V_\textrm{C}\). After the network reaches a relaxed state, these sources are disconnected from the network and the synaptic weight voltages are connected to the axons. The network then converges to the nearest trained pattern.

\begin{figure}
\centering
\includegraphics[width=\columnwidth]{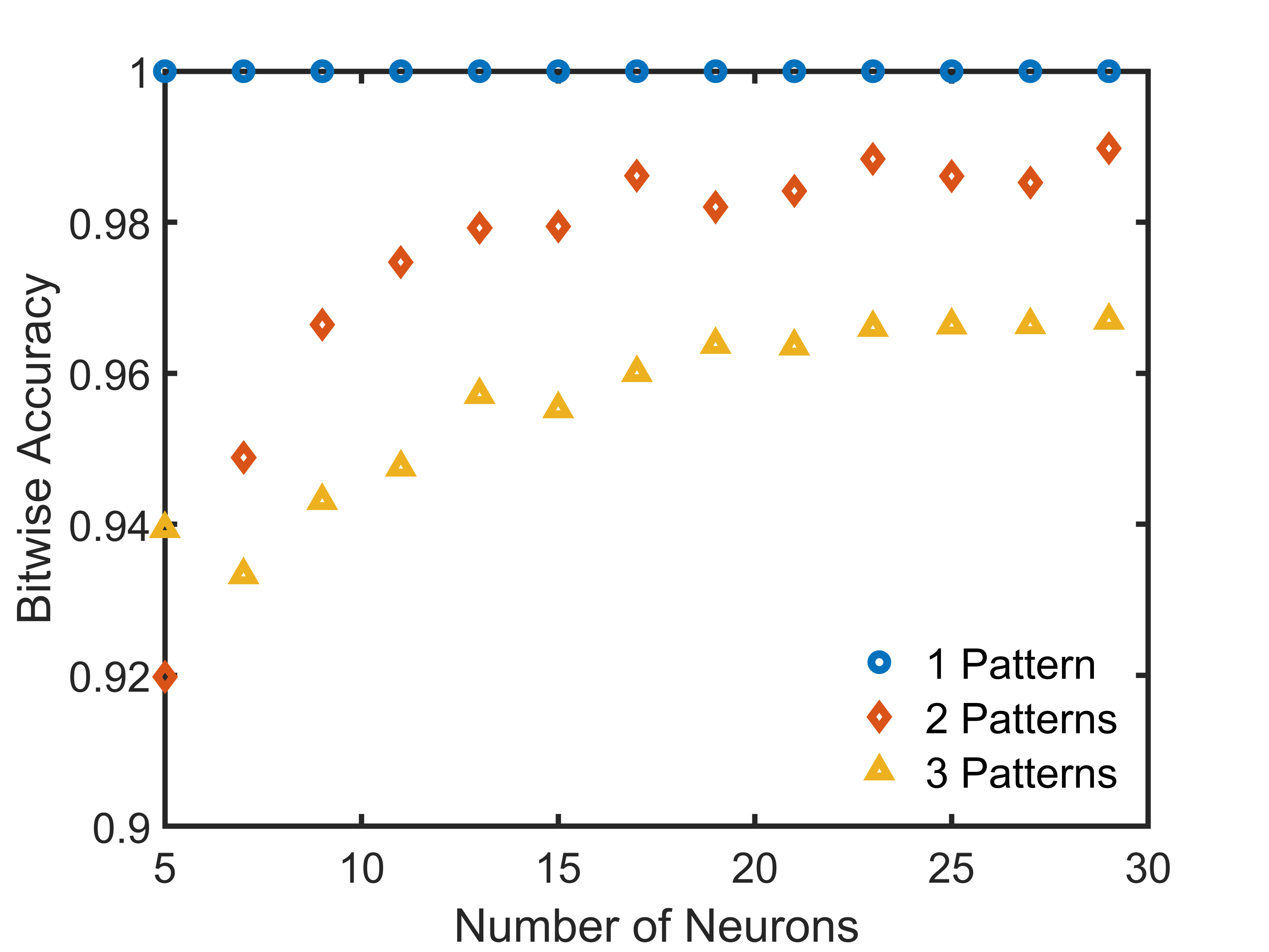}
\caption{Bitwise convergence accuracy for random pattern recall.}
\label{Figure 5}
\end{figure} 

\section{Computational Capabilities \& Efficiency}

To demonstrate the computational capabilities of this spintronic biomimetic RNN, three distinct tasks were carried out by this network: convergence to random patterns from random states, restoration of distorted image patterns, and determination of the max-cut division of a graph. In all three cases, the network successfully performed the tasks with accuracy and efficiency superior to the best previously-proposed RNNs. Our network was found to generate solutions quicker and with less energy than the alternatives technologies.  \par

\begin{table}[t]
    \centering
\caption{Device Parameters Used for Simulations}
\begin{center}
\begin{tabular}{ |c|c|c|c| } 
 \hline
 Symbol & Parameter & Value & Units \\ 
 \hline
\(g\) & Land\'e factor & \(2.1\) & - \\ 
\(P\) & Polarization constant & \(0.7\) & - \\ 
\(M_\textrm{SAT}\) & Saturation magnetization & \(8\times 10^{5}\) & A/m \\
\(A\) & Cross sectional area & \(50\) & \( \textrm{nm}^2 \) \\ 
Len & DW track length & \(100\) & nm \\ 
\(MTJW\) & MTJ width & 20 & nm \\ 
\(MTJP\) & MTJ placement & 50 & \% \\ 
\(L_\textrm{SYN}\) & Axon leak speed & \(-5\) & m/s \\ 
\(L_\textrm{AXON}\) & Soma leak speed & \(0.2\)& m/s \\ 
\(R_\textrm{P}\) & MTJ parallel resistance & \(500\)& \(\Omega\) \\ 
\(R_\textrm{AP}\) & MTJ antiparallel resistance & \(2000\)& \(\Omega\) \\ 
\(R_\textrm{M}\) & Metal layer resistance & \(2000\)& \(\Omega\) \\ 
\(V_\textrm{C}\) & Charge-up voltage  &\(0.1\)  &V  \\ 
\(W\) & Synaptic weight magnitude  &\(0.1\) &  V \\ 
\(V_\textrm{DW}\) & Soma to axon voltage  &Varies with N   &V \\ 
 \hline
\end{tabular}
\end{center}
\label{Table 1}
\end{table}

The RNNs were simulated in Cadence using a SPICE model of the DW-MTJ that incorporates its electric and magnetic behavior with the parameters listed in Table \ref{Table 1} \cite{Xiao2019EnergyAdder, Hu2020ProcessLogic, Hu2019SPICE-OnlyLogic}. In order to ensure that the axon mirrors the state of the soma, the value for \(V_\textrm{DW}\) was selected with the following design rule:\par
\begin{equation}
V_\textrm{DW}=-\frac{L_\textrm{SYN}R_\textrm{P}}{k}(N-1)(1+\frac{T}{2})
\label{vdw}
\end{equation} As network sizes larger than 20 neurons require prohibitively long simulation times, a simplified simulation technique was developed for those larger networks and validated against SPICE simulations.

\subsection{Computational Accuracy}

When evaluating the accuracy of an RNN, it is critical to note that the weight matrix of a Hopfield network is identical when storing a pattern or its inverse. For example, storing the pattern “110” or its inverse “001” results in identical Hopfield networks \cite{Hopfield1982NeuralAbilities.}. Therefore, in simulations, an output was considered accurate if it matched the desired pattern or its inverse. With this criteria and the parameters detailed in the appendix, our system was found to exhibit exceptional accuracy in both the recall of distorted memories and the calculation of approximate solutions to the max-cut problem.  \par

\subsubsection{Random Pattern Recall}

In this task, the Hopfield network is programmed to recall random patterns, and then provided with random input patterns; the network is considered successful if it converges to any one of the programmed patterns or their inverse. Networks of up to 29 neurons had their accuracy averaged across 1,000 test cases with a randomly selected pattern and input. Due to the number of input and test pattern pairs growing with $2^{2n}$, a random sampling test procedure was used to determine the accuracy across a large number of neurons.\par

As shown in the results of Fig. \ref{Figure 5}, the network achieved perfect accuracy for all single-pattern cases. Accurate recall was also achieved when two patterns were programmed into the network, eventually approaching 99\% bit accuracy. With three patterns, an average of more than 96\% of the bits in a pattern were resolved for all networks larger than 17 neurons.

\subsubsection{Distorted Image Recall}
\begin{figure}
     \centering
     \begin{subfigure}[b]{\columnwidth}
         \centering
         \includegraphics[width=\columnwidth]{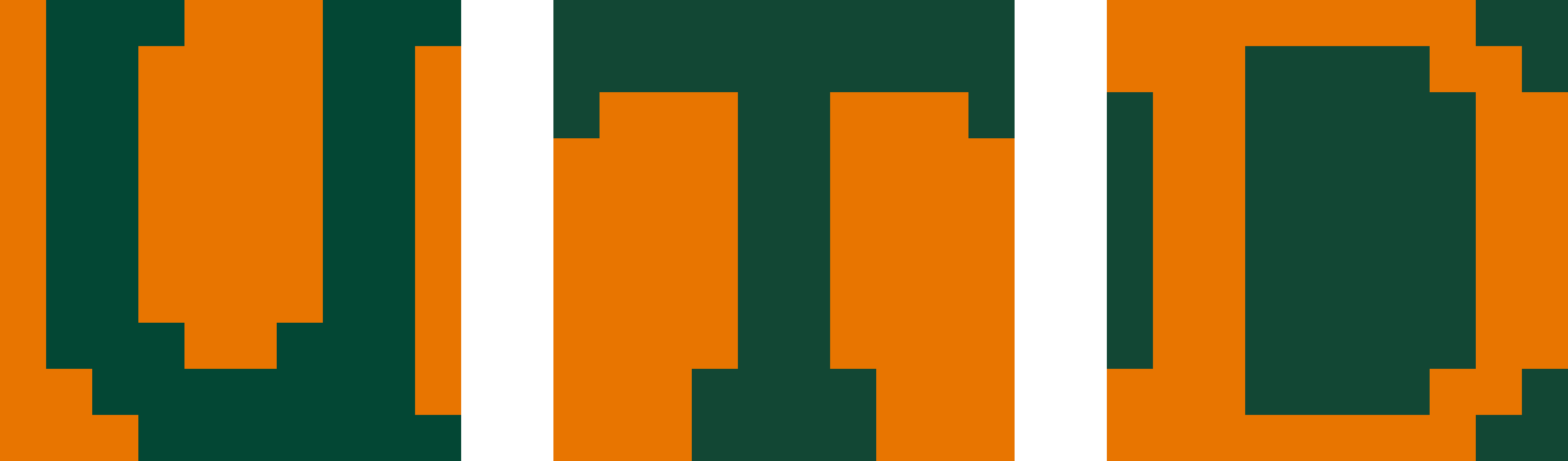}
         \caption{}
         \label{Figure 6a}
     \end{subfigure}
     \hfill
     \begin{subfigure}[b]{\columnwidth}
         \centering
         \includegraphics[width=\columnwidth]{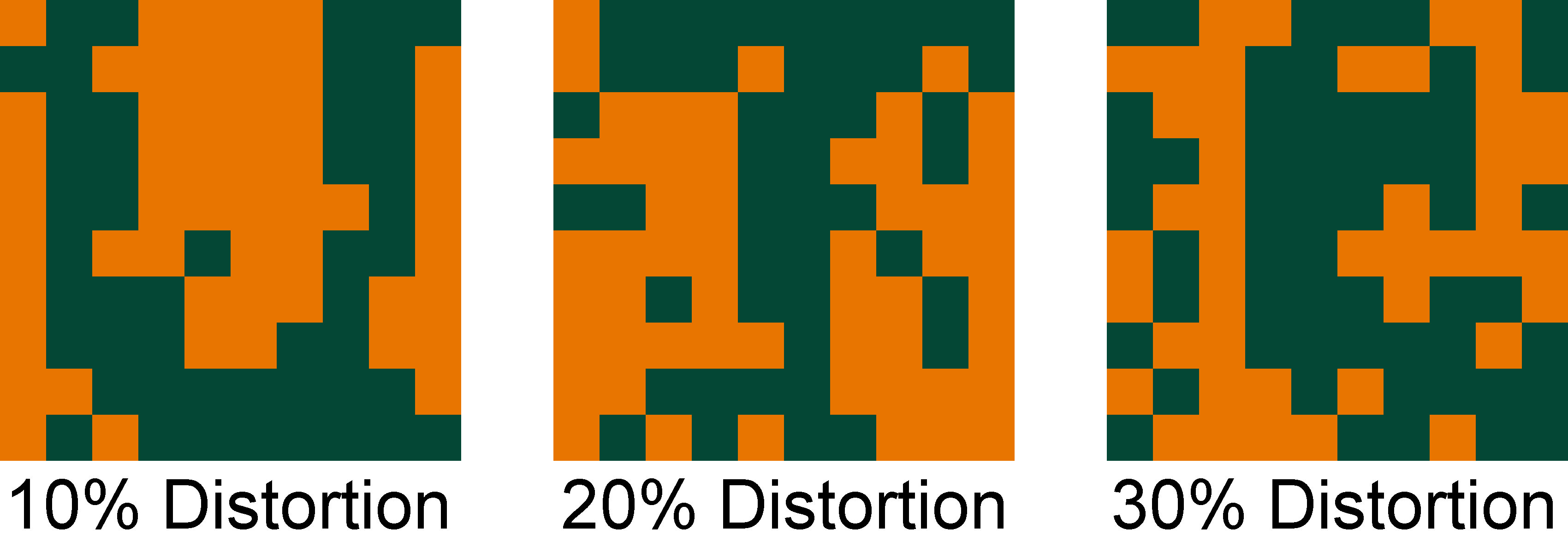}
         \caption{}
         \label{Figure 6b}
     \end{subfigure}
     \hfill
     \begin{subfigure}[b]{\columnwidth}
         \centering
         \includegraphics[width=\columnwidth]{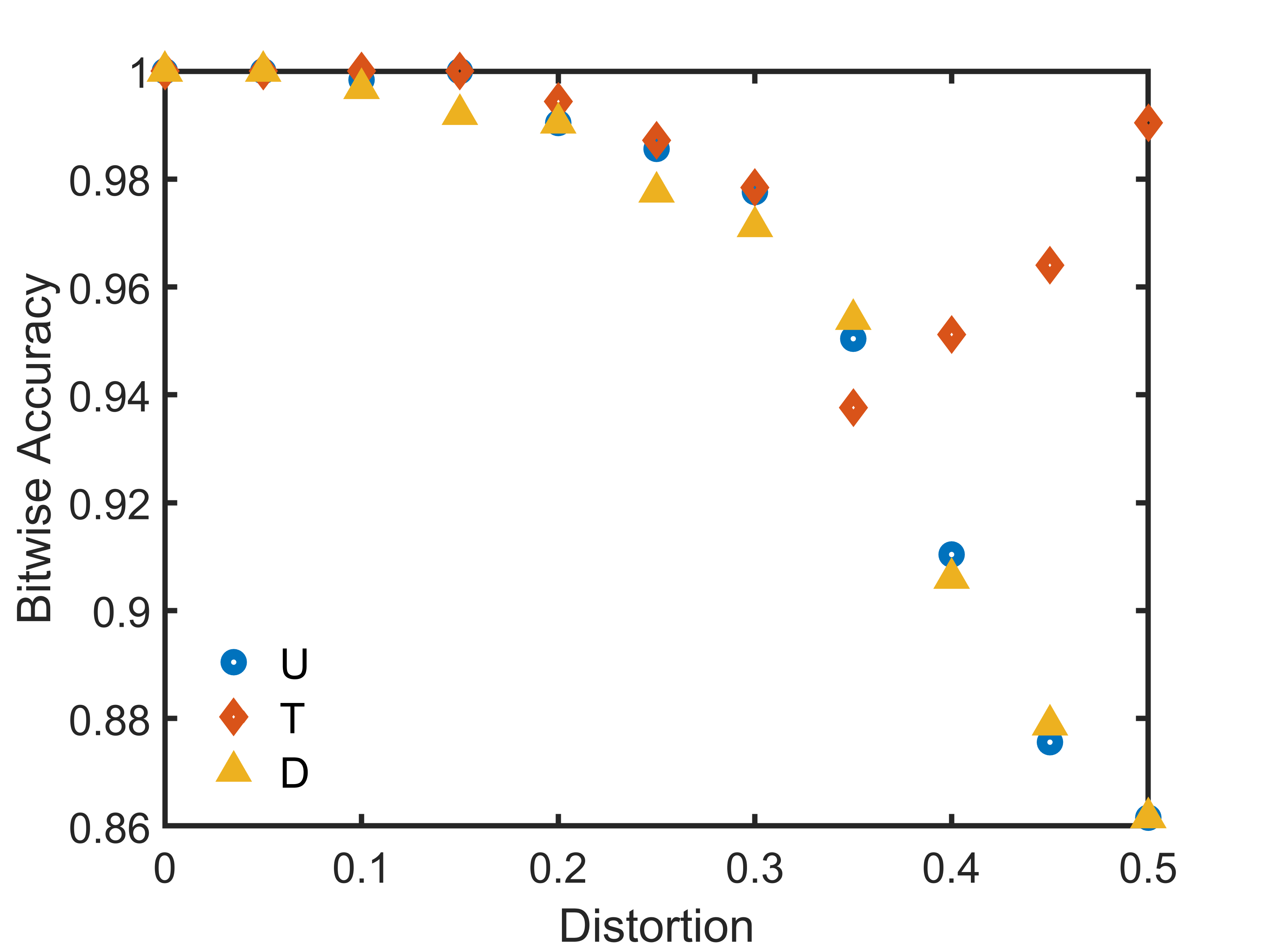}
         \caption{}
         \label{Figure 6c}
     \end{subfigure}
        \caption{(a) Three distinct 10x10 binary images programmed into a spintronic Hopfield network. (b) Sample inputs to the network with varying levels of distortion. (c) Bitwise recall accuracy as a function of distortion level.}
        \label{Figure 6}
\end{figure} 
One of the many uses of Hopfield networks is the reconstruction of images distorted by noise. As a test case, the three 10x10 binary images shown in Fig. \ref{Figure 6}(a) were programmed into a network. Each image was then subjected to distortion with varying percentages of flipped bits, as illustrated in Fig. \ref{Figure 6}(b). These distorted images were then fed into the network; the network was considered successful if it converged to the original image. A distortion of 0.5, or flipping half the bits, is the greatest possible amount of distortion, as inverse patterns are intrinsically programmed. \par

The average accuracy of the spintronic network as a function of distortion level is shown in Fig. \ref{Figure 6}(c). Perfect recall was achieved for low levels of distortion, and the spintronic network was able to converge to nearly-correct patterns even when provided with extremely high levels of distortion. At distortion levels of 0.33 or more, the pattern can become unrecognizable or closer to a different pattern, causing false convergences and lower accuracy. When distortion increases past 0.35, the ``T'' pattern accuracy increases while all other letters' accuracies decrease; due to the nature of this set of images, the network interprets noise as a ``T" pattern. 

\subsubsection{Max-Cut}
\begin{figure}
     \centering
         \includegraphics[width=\columnwidth]{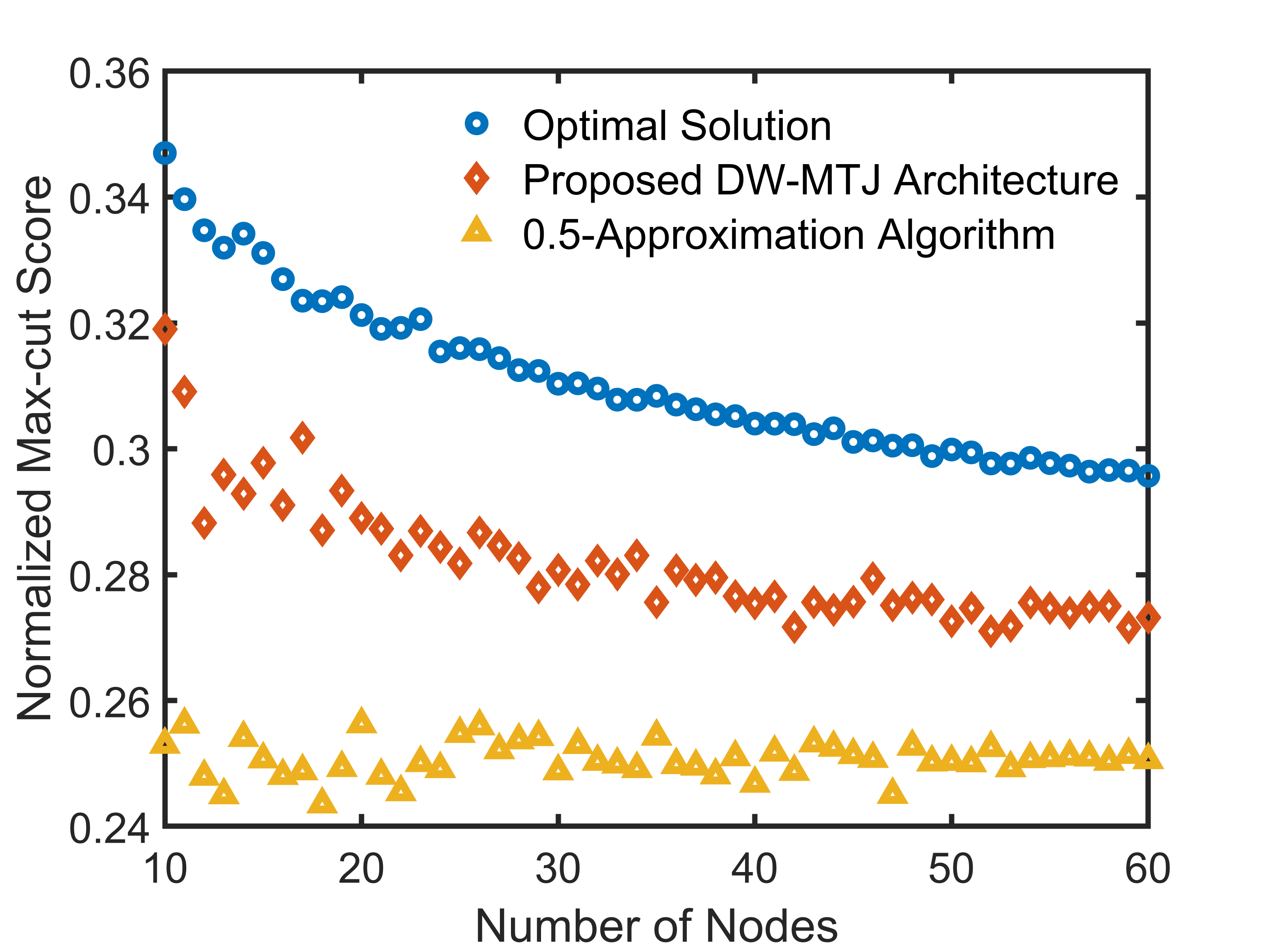}
        \caption{Normalized score of the optimal solution, solution generated by a 0.5-approximation algorithm, and a solution generated by the proposed DW-MTJ architecture. An average based on an input of 20 of the same 50\%-connected randomly generated graphs were used for each data point.} 
        \label{Figure 6.5}
\end{figure} 

 The max-cut problem is a well known NP-complete problem in which the nodes of a simple, undirected graph are divided into two sets such that the weights of the edges between the sets are maximized \cite{Wu2001}. It is often useful to provide a fast approximate solution \cite{Goemans1995}.\par 
Our biomimetic RNN architecture was used to find approximate max-cut solutions for undirected graphs. Each neuron corresponds to a graph node, such that a network requires the same number of neurons as nodes in the graph. The weights between neurons representing disconnected nodes were set to \(+W\), while weights between connected nodes were set to \(-1.05W\); this 5\% difference was chosen empirically to introduce asymmetry and maximize accuracy. The initial states of all the neurons' DW-MTJs were set to the anti-parallel state before release. After network convergence, the set to which each neuron belongs is represented by the state of its DW-MTJ. \par

Fig. \ref{Figure 6.5} shows the average max-cut score of our algorithm normalized by the total number of connections for a 50\% connected graph of varying sizes. This is compared against the optimal solution and that of a naive 0.5-approximation algorithm. Across all sizes, our network greatly outperformed the 0.5-approximation algorithm, and averaged a 9.02\% difference from the optimal solution. 

\subsection{Computational Efficiency}

In addition to accuracy, the efficiency of a computing system is a critical metric of its utility. While area is important, this study focuses on the time-to-solution, power dissipation, and energy consumption.

The time-to-solution is calculated as the time between when the network starts to load the initial state and when all of its DWs are stationary post-convergence. We have approximated the energy consumption of the network based on the assumption that half of its MTJs are in the parallel state and half are in the anti-parallel state; all power calculations are then performed by dividing the total energy use by the total run time across all trials.

\subsubsection{Random Pattern Recall}

The single pattern associative memory task achieved a fast time-to-solution: over 100 trials, a 60-neuron spintronic Hopfield network reached an approximate solution after an average of 81.2 ns, with no trial taking longer than 113 ns to converge. The approximate average power dissipation of the network was 44.3 mW, and the average energy consumption was 3.60 nJ; the worst-case energy consumption was 5.23 nJ.\par

\subsubsection{Distorted Image Recall}

While similar to the random pattern recall procedure, the binary image distortion task required a 100-neuron spintronic Hopfield network. While the convergence time was slightly faster, the larger network resulted in greater energy consumption. Over 1000 trials restoring the "U", "T" and "D" images with the varying distortion levels shown in Fig. \ref{Figure 6}, the network consumed an average of 11.04 nJ over 78.14 ns. \par

\subsubsection{Max-Cut}

For max-cut simulations, ten standard 50\% connected graphs in the Biq Mac Library \cite{Wiegele2007BiqLibrary} were used as inputs to a 60-neuron spintronic Hopfield network. Across all runs, the spintronic network found a solution after an average of 344 ns. The spintronic network also consumed minimal energy, with an average of 50.6 mW of power dissipation and 17.4 nJ of energy consumption. This task took longer to converge compared to the memory-based ones, due primarily to the uniformly anti-parallel starting state of the neurons: the smaller currents associated with the anti-parallel state result in slower DW motion until the neurons begin to flip into the parallel state. 

\section{Comparison \& Optimization}

To evaluate its utility in neural network applications, this spintronic RNN was compared to alternative architectures, leading to the conclusion that the spintronic RNN reaches a solution of similar quality in less time, with less power, and with less energy. Furthermore, we analyzed the impact of the synaptic weight voltage on energy consumption and convergence time, enabling system optimization for maximum efficiency.

\subsection{Comparison to Alternative Approaches}

\begin{table}[t]
    \centering
\caption{Max-Cut Efficiency Comparison}
\begin{center}
\begin{tabular}{ |c|c|c|c| } 
 \hline
 Work & Time-to-Solution (ns) & Power (mW) & Energy (nJ) \\ 
 \hline
Memristor RNN & 600 & 120 & 72 \\ 
SONOS RNN & Unknown & Unknown & 33 \\ 
Spintronic RNN & 344 & 50.6 & 17.4 \\ 
 \hline
\end{tabular}
\end{center}
\label{Table 0}
\end{table}

A recent study \cite{Cai2020Power-efficientNetworks} described a Hopfield network that used memristors to find solutions to the max-cut problem (alongside a required co-processor, presumably requiring CMOS). In this work, a switch matrix and driver circuit was used to select rows of a memristor array to perform multiplication. The result was then filtered to produce a threshold function, and then fed back into the input/output buffers to be loaded in during the next cycle. They estimate that for a network of 60 neurons with 50\% connectivity, 50 cycles of the device would require 72 nJ over 600 ns. These values correspond to a probability of converging to the correct solution of between 30\% and 40\%. \par

In contrast, our spintronic network does not include cycles, and is composed solely of DW-MTJ devices and voltage sources. This allows for a faster convergence time and less energy use. Compared to the quicker of their two configurations of the memristive Hopfield network, the spintronic network converged 74.4\% faster. The spintronic network also uses less energy, consuming 24.4\% less energy. The spintronic network's worst-case energy consumption, out of the ten trials, is an outlier at 47.6 nJ, but still lower than the memristive network's. 

\par
Another work has modeled a Hopfield network based on memristive SONOS transistors \cite{YiARXIV2021}. For the same Biq Mac library and network size, their energy consumption was 33 nJ, which is 89.7\% greater than our proposed spintronic network. Their work also requires external circuitry to read the output of each cycle in order to feed it as input to the next cycle. While their work did not specify the cycle time, it mentions a SONOS ramp-up time of 1 ms for a different demonstration. We therefore suspect that our network is significantly faster. 
\par

\subsection{Network Optimization}
\begin{figure}
     \centering
         \includegraphics[width=\columnwidth]{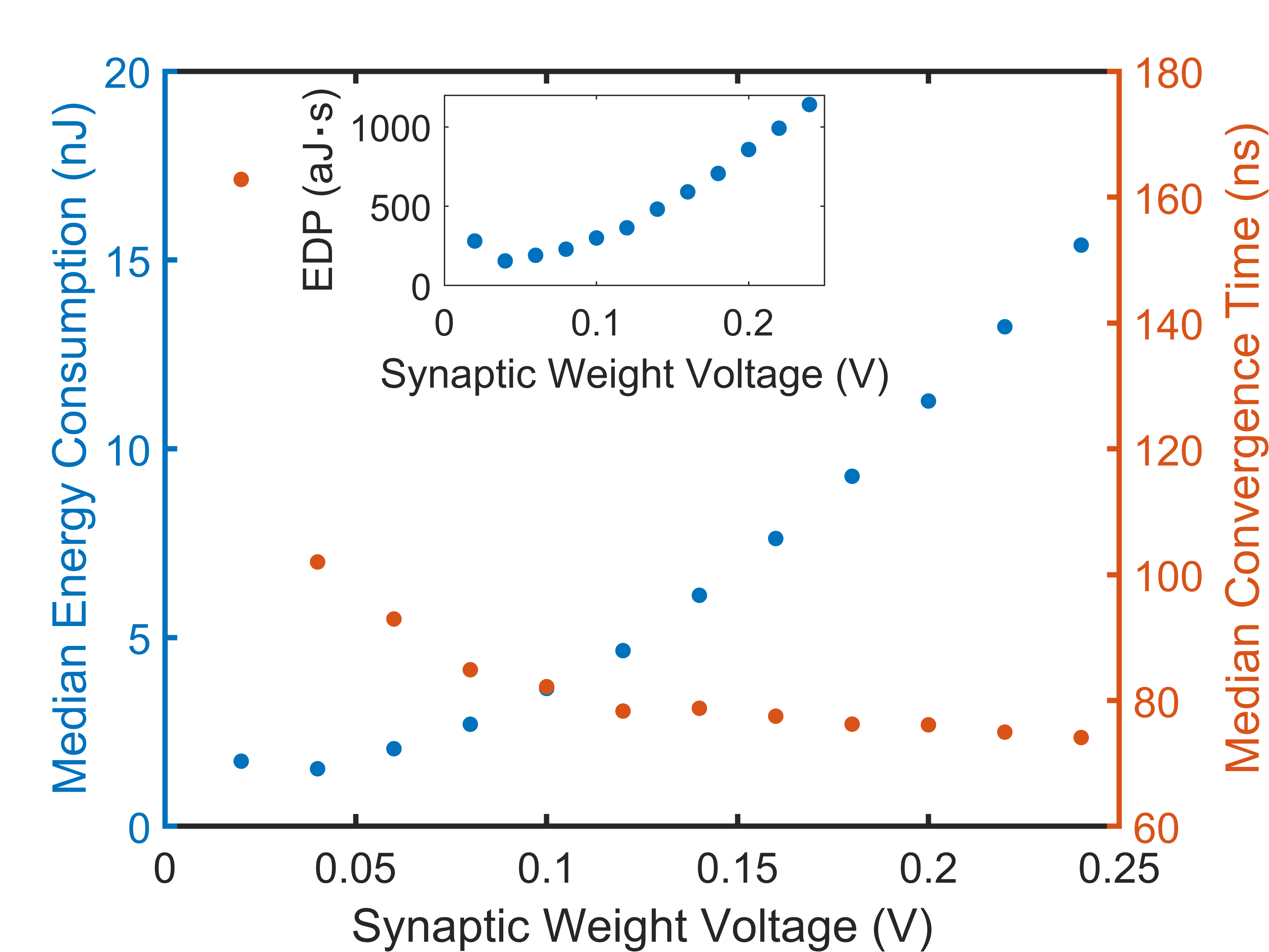}
         \label{Figure 7a}
        \caption{ Approximate median convergence time and energy consumption of a 60-neuron network with $V_\textrm{C}=0.25V$ as a function of synaptic weight voltage $W$. Inset: The energy-delay product (EDP) for this network.}
        \label{Figure 7}
\end{figure} 

The efficiency of the spintronic network depends on both device parameters and voltage levels. Device variables such as the MTJ dimensions, metal track resistance, and TMR depend on manufacturing technology and are difficult to tune. On the other hand, voltage levels in the inputs and weights are easily modified. The synaptic weight magnitude  levels can drastically affect power consumption, as illustrated in Fig. \ref{Figure 7}. In general, increased synaptic weight voltages benefit the network by increasing convergence speed but require greater power consumption.
For the parameters tested, the minimum energy-delay product (EDP) is at approximately $W=0.04$ V with a value of 0.16 aJ$\cdot$s.

\FloatBarrier

\section{Conclusion}
This paper presents a biologically inspired, asynchronous, fully spintronic recurrent neural network built solely from DW-MTJs. Neurons are cascaded into a Hopfield network, demonstrating the first biomimetic, CMOS-free RNN with reconfigurable weights. The network successfully performed associative memory and max-cut problems in SPICE simulations, while converging faster and requiring less energy consumption than previously-proposed approaches. This spintronic RNN thus makes effective use of biomimicry with spintronic devices to provide a promising approach for artificially intelligent systems with exceptional efficiency.

\section*{Acknowledgment}
The authors thank R. Iyer for editing, E. Laws, J. McConnell, N. Nazir, L. Philoon, and C. Simmons for technical support, and H.-J. Drouhin for arranging the research exchange. This research is sponsored in part by the National Science Foundation under CCF award 1910800 and the Texas Analog Center of Excellence undergraduate internship program. 

\FloatBarrier


\end{document}